\documentclass[10pt,twocolumn,letterwork]{article}

\usepackage{cvpr}              
\usepackage[accsupp]{axessibility}
\usepackage{tabularx}
\usepackage{multirow}
\usepackage{graphicx}
\usepackage{algorithm}
\usepackage{algorithmic}

\usepackage[table]{xcolor}

\definecolor{cvprblue}{rgb}{0.21,0.49,0.74}
\usepackage[breaklinks,colorlinks,allcolors=cvprblue]{hyperref}

\def\paperID{} 
\def\confName{CVPR}
\def\confYear{2026}

\title{Delving Aleatoric Uncertainty in Medical Image Segmentation via Vision Foundation Models}

\author{Ruiyang Li, Fang Liu\thanks{Corresponding Author: Fang Liu (This is the author’s version of a paper accepted to CVPR 2026)}, Licheng Jiao, Xinglin Xie, Jiayao Hao, Shuo Li, \\
    Xu Liu, Jingyi Yang, Lingling Li, Puhua Chen, Wenping Ma \\
Key Laboratory of Intelligent Perception and Image Understanding of Ministry of Education,\\
International Research Center for Intelligent Perception and Computation,\\
Joint International Research Laboratory of Intelligent Perception and Computation,\\
School of Artificial Intelligence, Xidian University, Xi’an 710071, China\\
{\tt\small liruiyang@stu.xidian.edu.cn; f63liu@163.com}
}

\begin{document}
\maketitle



\begin{abstract}

Medical image segmentation supports clinical workflows by precisely delineating anatomical structures and lesions. However, medical image datasets medical image datasets suffer from acquisition noise and annotation ambiguity, causing pervasive data uncertainty that substantially undermines model robustness. Existing research focuses primarily on model architectural improvements and predictive reliability estimation, while systematic exploration of the intrinsic data uncertainty remains insufficient. To address this gap, this work proposes leveraging the universal representation capabilities of visual foundation models to estimate inherent data uncertainty. Specifically, we analyze the feature diversity of the model's decoded representations and quantify their singular value energy to define the semantic perception scale for each class, thereby measuring sample difficulty and aleatoric uncertainty. Based on this foundation, we design two uncertainty-driven application strategies: (1) the aleatoric uncertainty-aware data filtering mechanism to eliminate potentially noisy samples and enhance model learning quality; (2) the dynamic uncertainty-aware optimization strategy that adaptively adjusts class-specific loss weights during training based on the semantic perception scale, combined with a label denoising mechanism to improve training stability. Experimental results on five public datasets encompassing CT and MRI modalities and involving multi-organ and tumor segmentation tasks demonstrate that our method achieves significant and robust performance improvements across various mainstream network architectures, revealing the broad application potential of aleatoric uncertainty in medical image understanding and segmentation tasks. 
\href{https://github.com/Lry777/CVPR_AUV}{The code is available.}
\end{abstract}

\section{Introduction}
\label{sec:intro}

\begin{figure}[tbp]
\centering
\includegraphics[width=1.0\linewidth]{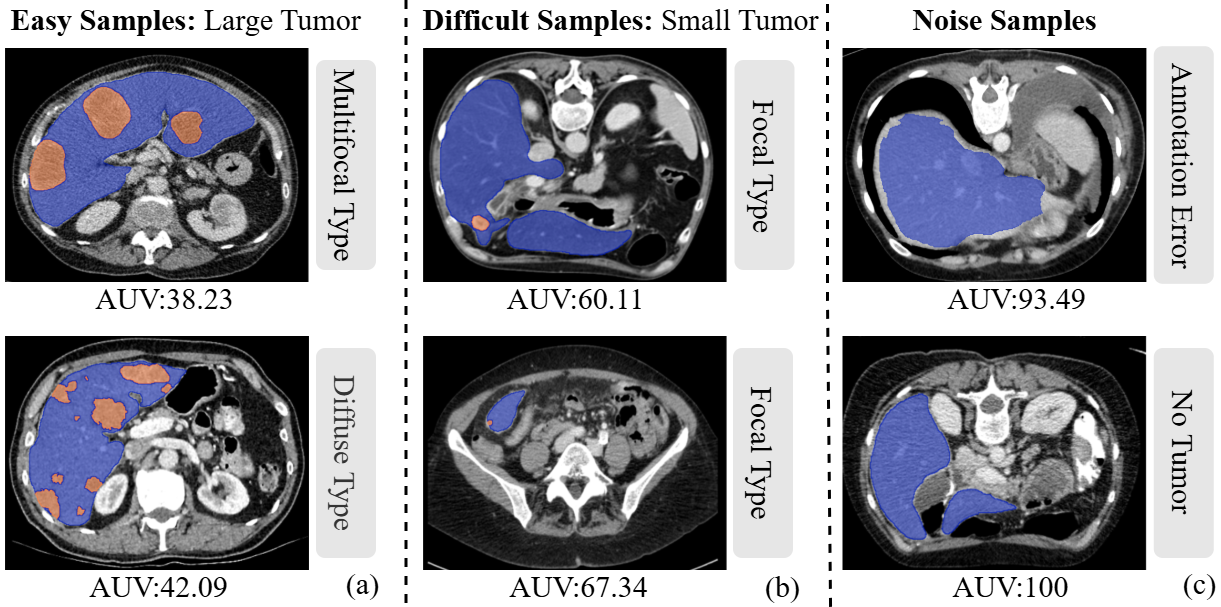}
\caption{Visualization of low (high) difficulty training samples scored by MedSAM2 via derived Aleatoric Uncertainty Values (AUV). Lower AUV indicates easier-to-learn samples.}
\label{motivation}
\end{figure}


Medical image segmentation aims to delineate clinically significant anatomical structures or specific regions from medical images, providing objective evidence for physician decision making in disease diagnosis, surgical navigation, and treatment planning \cite{moor2023foundation, liu2025biomedical, jiao2026foundation}. However, the acquisition of medical imaging data faces numerous challenges, including multi-center device discrepancies, imaging noise interference, and the reliance on expert knowledge with high annotation costs \cite{yang2025structural}. These factors inevitably introduce randomness and noise into medical datasets, resulting in inherent data uncertainty (aleatoric uncertainty). If deep learning-based medical image segmentation methods learn directly from these uncertain or noisy samples, they are prone to misleading feedback or overconfidence. Therefore, quantifying such uncertainty is crucial for understanding the intrinsic characteristics of the data and developing more robust segmentation models.


Other researches demonstrate that quantifying aleatoric uncertainty can not only identify inherent noise in data, but also effectively reflect the varying learning difficulties between samples, thereby guiding the optimization of model training \cite{paul2021deep, sorscher2022beyond, peng2022angular, kirchhof2023url, attia2025data}. In medical image segmentation tasks, different clinical scenarios often present distinct challenges. For example, organ and lesion boundaries frequently exhibit ambiguity and high morphological heterogeneity; significant disparities exist in intensity distributions between imaging modalities; and small targets or low-contrast structures (e.g., tumors and lesions) often suffer from low discernibility \cite{isensee2021nnu, yang2023ept, li2023dht, yan20233d}. These factors indicate that, beyond data noise, medical images inherently possess varying levels of difficulty in both clinical interpretation and algorithmic learning. However, current research on uncertainty in medical imaging focuses primarily on the reliability of model predictions \cite{huang2024review, yang2025structural}, with limited attention dedicated to the inherent aleatoric uncertainty of the data itself. Therefore, this work aims to investigate aleatoric uncertainty in medical image segmentation tasks and explore its practical implications in specific clinical contexts.



Due to constraints in temporal and resource costs, manual assessment and annotation of inherent aleatoric uncertainty in medical images present significant challenges. Existing studies typically evaluate sample-level aleatoric uncertainty by analyzing task-specific data distributions and leveraging feature vectors from trained models \cite{cui2023learning, zhang2024one, cui2024exploring, mucsanyi2024benchmarking}. However, models trained on specific task datasets are prone to overfitting, and their performance often relies on manually designed strategies, which may compromise the reliability of uncertainty estimation. Visual foundation models, pre-trained on large-scale datasets, can learn highly discriminative structured feature representations and demonstrate exceptional capabilities in generalization, robustness and zero-shot transferability \cite{liu2023clip, du2024segvol, ma2025medsam2,awais2025foundation}. These models map multi-source heterogeneous data into a unified and stable task-agnostic feature space, substantially improving the accuracy of data uncertainty quantification \cite{cui2023learning}. Based on this, we propose an innovative application that leverages feature vectors extracted from medical visual foundation models to precisely quantify aleatoric uncertainty in medical imaging. Specifically, by leveraging the structural complexity of the feature vectors decoded by the foundation model, we assess sample difficulty through calculating the singular value energy distribution of the vectors to quantify feature diversity, and define this metric as the semantic perception scale. Samples with characteristic discriminative features contain more diverse structural information (easy samples in Figure \ref{motivation}-a), resulting in a more diverse energy distribution in the singular value spectrum of their feature vectors, which corresponds to lower uncertainty. In contrast, samples with ambiguous or indistinct features demonstrate sparser singular value distributions (noise samples in Figure \ref{motivation}-c), reflecting higher uncertainty measures. Finally, by computing the Semantic Uncertainty Scale for all classes within each sample and applying global normalization, we derive the Aleatoric Uncertainty Value (AUV) for all samples, as shown in Figure \ref{motivation}.


Building upon the aforementioned theoretical foundation, we propose two practical and critical application strategies of aleatoric uncertainty in medical image segmentation tasks: aleatoric uncertainty-aware data filtering and dynamic uncertainty-aware optimization strategy. First, based on medical imaging characteristics and task requirements, we define multiple data filtering strategies using the aleatoric uncertainty value to eliminate noisy samples that may misguide model training, thereby enhancing learning efficiency. In addition, our study reveals a high consistency between the model's semantic perception scale for each class and its prediction scores, which operate without relying on annotations (see Table \ref{Correlation} for details). Based on this finding, we propose a plug-and-play adaptive loss function that dynamically adjusts class-specific weights in the loss function using the semantic perception scale as a regularization term, combined with a learnable label denoising mechanism to improve training stability.


The aleatoric uncertainty value serves as an additional annotation for training data and can be deployed in a plug-and-play manner across diverse segmentation scenarios and models. We conduct extensive experiments on five challenging benchmark datasets that encompass multi-organ and tumor segmentation tasks using both CT and MRI modalities. Experiments demonstrate that discarding noisy samples based on aleatoric uncertainty value yields significant performance improvements for baseline networks, while leveraging the generalization capabilities of medical visual foundation models surpasses other quantification strategies. Furthermore, the dynamic uncertainty-aware optimization strategy improves model robustness against both noisy samples and annotation artifacts, with its effectiveness extensively validated on advanced baseline networks incorporating CNN, Transformer, and Mamba architectures. Finally, detailed ablation studies confirm the robustness of hyperparameter configurations while further exploring the potential of aleatoric uncertainty quantification. In summary, this work makes the contributions as follows:

\begin{itemize}


\item [$\bullet$] We first introduce large-scale pre-trained visual foundation models into aleatoric uncertainty estimation for medical imaging. By constructing a semantic perception scale through the singular value energy distribution of feature matrices, we achieve accurate measurement of sample difficulty and uncertainty without relying on annotation.


\item [$\bullet$] We propose an aleatoric uncertainty-aware data filtering mechanism to eliminate potentially noisy samples and improve training data quality. Additionally, we introduce a dynamic uncertainty-aware optimization strategy that enhances baseline model performance and robustness through adaptive loss weighting and label denoising.


\item [$\bullet$] Comprehensive validation on five challenging segmentation datasets demonstrates the significant effectiveness and broad applicability of our proposed method in enhancing model robustness and segmentation performance, exploring the potential of aleatoric uncertainty in medical image segmentation.

\end{itemize}

\section{Aleatoric Uncertainty Analysis}
\label{sec:SUS}



This section aims to quantify aleatoric uncertainty in medical image segmentation, which arises from inherent ambiguities in images, noise, or annotation inconsistencies. Unlike epistemic uncertainty that characterizes model limitations, data uncertainty reflects intrinsic randomness in the task itself, necessitating an efficient measurement approach independent of multiple model inferences or Monte Carlo sampling \cite{ji2023uncertainty}. Inspired by brain cognition mechanisms \cite{ma2022delving, von2023recurrent,jiao2024ai, jiao2021deep,jiao2022new,liu2023bio,jiao2024multiscale}, we propose to evaluate the uncertainty of medical images based on the subjective perception of medical vision foundation models. To this end, from perspectives of matrix theory and information theory, we quantify the feature diversity of images in the semantic space to represent the model's subjective perception, which we define as semantic perception scale. Finally, through global normalization, the Aleatoric Uncertainty Value (AUV) is measured for each sample. More details about the related work are provided in the Appendix.

\subsection{Feature extraction and energy distribution}


Existing studies have shown that aleatoric uncertainty is reflected in the complexity of the model's feature space. \cite{zhang2024one}. Therefore, we posit that in medical images, features extracted from highly uncertain categories (e.g., those with blurred boundaries, atypical pathology, or artifacts) tend to exhibit more disorganized and sparse patterns, whereas features from highly certain categories demonstrate richer and more compact patterns \cite{li2022unsupervised, li2023minent}. To capture this structural characteristic, we propose using a large-scale pre-trained medical foundation model as a fixed feature extractor $V(\cdot, \theta)$. Given an input image volume $X \in \mathbb{R}^{D \times H \times W}$, we feed it into $V(\cdot, \theta)$ to obtain the feature vectors $Z$ corresponding to the image: 

\begin{equation}
\begin{aligned}
& Z = V(X, \theta) \in \mathbb{R}^{C \times D \times H \times W},
\end{aligned}
\end{equation}

\noindent where, $C$ denotes the number of semantic channels, $D, H, W$ represents the spatial dimensions, and $Z_i = [z_1,z_2...,z_c]$ encodes the rich semantic information of the $i$-th input image, and $c$ represents the specific class.


Inspired by the application of multivariate Gaussian distributions in classification tasks, the covariance matrix is employed to characterize the shape and orientation of within-class feature distributions \cite{cui2023learning}. To analyze the 3D spatial feature maps of specific class, we reshape them into a 2D matrix $z_c \in \mathbb{R}^{D \times (H \cdot W)}$, expressed as follows:

\begin{equation}
\begin{aligned}
& \mathbf{\Sigma} = \frac{1}{N - 1} ( z_c - \bar{z_c} ) ( z_c - \bar{z_c})^{\mathrm{T}},
\end{aligned}
\end{equation}


\noindent where $\bar{z_c}$ is the mean vector of $z_c$. The magnitude of the eigenvalues $\lambda_j$ of $\mathbf{\Sigma}$ quantitatively describes the variance (degree of uncertainty) of feature vectors along different orthogonal directions. A large eigenvalue indicates significant dispersion of the data along the corresponding eigenvector direction, representing an important mode of variation. To bypass the explicit computation of the covariance matrix, we perform singular value decomposition (SVD) on $z_c = US_cV^T$, where the diagonal elements $\sigma_1^c \geq \sigma_2^c \geq ... \geq \sigma_r^c \geq 0$ of $S$ are singular values. A profound and direct connection exists between SVD and the eigendecomposition of the covariance matrix: ${z_c}^{T} z_c = (V S_c U^{T})(U S_c V^{T}) = V S_c^{2} V^{T}$, which means that the eigenvalues of the covariance matrix are exactly equal to the squares of the singular values, $\lambda_{j}^c = (\sigma_{j}^c)^2$. Consequently, the singular value spectrum $\mathbf{\sigma}^c = (\sigma_1^c, \sigma_2^c, ..., \sigma_r^c)$ from SVD provides information equivalent to that obtained from eigendecomposition, but is computationally more direct and stable.


Unlike methods directly based on multivariate Gaussian density estimation, we further fit the energy distribution of singular values in the feature space and define this as the contribution of each mode to the total variance. The total variance of the feature matrix $z_c$ is given by the square of the Frobenius norm, which equals the sum of the squares of all singular values:

\begin{equation}
\begin{aligned}
& \|z_c\|_{F}^{2} 
= \sum_{i,j} ({z^c}_{ij})^{2} 
= \sum_{j=1}^{r} (\sigma^c_{j})^{2} 
= \sum_{j=1}^{r} \lambda^c_{j}.
\end{aligned}
\end{equation}


\noindent Thus, we naturally define the normalized squared singular value spectrum as the intrinsic energy distribution $p$ of class $c$ under a specific analytical perspective:

\begin{equation}
\begin{aligned}
&  p_{j}(z_c) = 
\frac{(\sigma^c_{j})^{2}}{\sum_{j=1}^{r} (\sigma^c_{j})^{2}+\varepsilon} 
= \frac{\lambda^c_{j}}{\sum_{j=1}^{r} \lambda^c_{j} + \varepsilon} 
, j=1,2,\ldots,r.
\end{aligned}
\end{equation}


\noindent where $\varepsilon$ is a smoothing parameter, and the distribution satisfies the normalization condition $\sum_{j=1}^r p_j(z_c) = 1$.

\begin{figure}[tbp]
\centering
\includegraphics[width=0.99\linewidth]{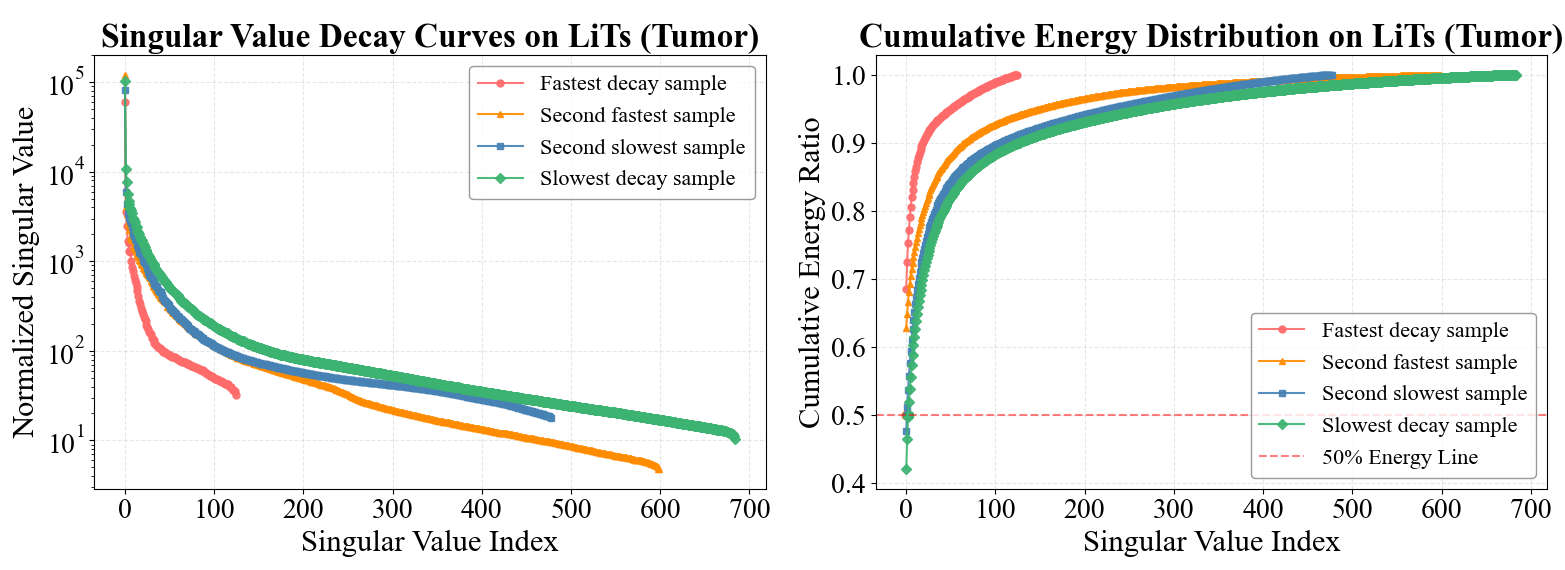}
\caption{Singular value decay (left) and cumulative energy (right), illustrating the disparities in feature distribution across four samples of varying difficulty from the LiTS (Tumor) dataset.}
\label{singular}
\vskip -0.1in
\end{figure}

\begin{figure*}[tbp]
\centering
\includegraphics[width=\textwidth]{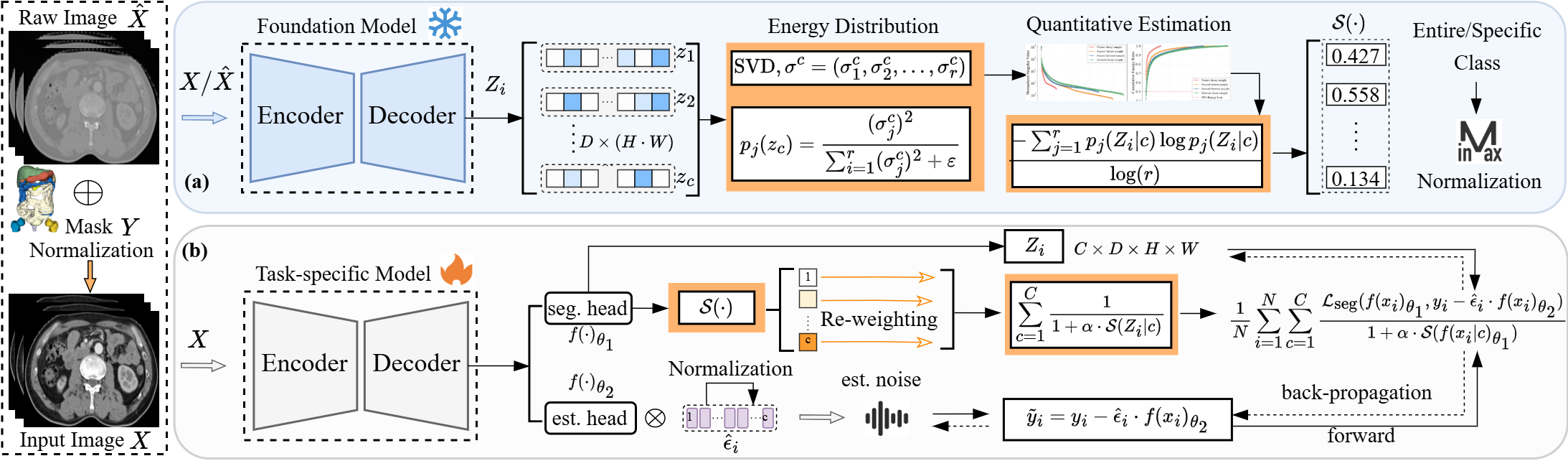}
\vskip -0.05in
\caption{Illustrative pipeline of the quantifying aleatoric uncertainty (orange part in (a)). Two applications of aleatoric uncertainty study in medical image segmentation tasks: aleatoric uncertainty-aware data filter (a) and dynamic uncertainty-aware optimization strategy (b).}
\label{method}
\vskip -0.2in
\end{figure*}

\subsection{Quantification of Aleatoric Uncertainty Value}


Based on the intrinsic energy distribution of the feature vectors described above, we measure the feature diversity of each sample in the training set by calculating its information content in the energy distributions of all classes \cite{ma2022delving, ma2025predicting, ma2024unveiling}. The decay rate of the singular values of the feature vectors reveals their intrinsic rank and redundancy: rapid decay indicates a low-rank structure, leading to rapid energy accumulation (red curve in Figure \ref{singular}), in contrast, slower decay indicates a full-rank structure, corresponding to a richer and more robust feature space with increased noise resistance, where the energy accumulation rate grows more gradually (green curve in Figure \ref{singular}).Specifically, we compute the Shannon entropy of the energy distribution to quantify its dispersion, followed by linear normalization:

\begin{equation}
\begin{aligned}
& \mathcal{S}(Z_i|c)=\frac{-\sum_{j=1}^r p_{j}(Z_i|c) \log p_{j}(Z_i|c)}{\log (r)},
\end{aligned}
\end{equation}


\noindent where, $\mathcal{S}(Z_i|c)$ denotes the semantic perception scale for the $c$-th class of the $i$-th image in the training data, while the overall uncertainty for the entire image satisfies $\mathcal{S}(Z_i) =\sum_{c=1}^C \mathcal{S}(Z_i|c)$. A fuller rank of $\mathcal{S}$ indicates richer feature diversity within the class, corresponding to higher values and lower aleatoric uncertainty. The detailed derivation of the aforementioned formulation is provided in the Appendix. Finally, to improve scale discriminability, we further apply a normalization technique combining logarithmic transformation with min-max scaling, mapping the Aleatoric Uncertainty Value (AUV) to the [0,1] interval:

\begin{equation}
\begin{aligned}
& \text{AUV}(Z_i) = 1-
\frac{\text{log}(\mathcal{S}(Z_i)) - \min \{\text{log}\mathcal{S}(Z_i)\}}
{\max \{\text{log}\mathcal{S}(Z_i)\} - \min \{\text{log}\mathcal{S}(Z_i)\}} .
\end{aligned}
\end{equation}


\noindent Here, $\text{AUV}(Z_i) \in (0, 1)$ represents the AUV of $Z_i$, where $\mathcal{S}(\cdot)$ can be flexibly selected as the semantic perception scale of all classes or specific classes. AUV closer to 1 indicates a higher predictive uncertainty for the corresponding class, while values closer to 0 signify a higher certainty.




\section{Disentangling Uncertainties in Medical Image Segmentation}

\subsection{Problem formulation}


This section begins with the medical segmentation task. The raw image is denoted as $\hat{X}$, $X$ is obtained by normalizing $\hat{X}$ according to the annotated mask regions, as shown in Figure \ref{method}. For the image input, there is a foundation network to map the input to the prediction $Z$. Considering the input feature distribution $x:f(x)$, a finite dataset $\mathcal{D}={(x_i,y_i)}_{i=1}^N$ represents $N$ inputs sampled from the input distribution along with their corresponding outputs. In real medical imaging data, noise is commonly present, indicating the presence of aleatoric uncertainty, which can be expressed as $\tilde{x}_i = x_i + \xi_i$, $\tilde{y}_i = y_i + \epsilon_i$, where ${\xi_i, \epsilon_i}$ represents two independent noise variables. This study aims to quantitatively estimate the aleatoric uncertainty in training data and explores its practical applications in medical image segmentation. We propose multiple AUV-based data filtering strategies designed to eliminate potential noisy and interference objects, enabling more efficient and reliable model training. Furthermore, we introduce a dynamic uncertainty-aware optimization strategy that leverages semantic perception scale and label noise estimation to enhance the model's learning efficiency. In particular, both proposed schemes achieve these improvements without introducing additional computational overhead during training and offer plug-and-play functionality.

\subsection{Aleatoric Uncertainty-aware Data Filtering}


We propose to leverage the statistical distribution characteristics of the aleatoric uncertainty value to identify and subsequently remove potential noisy samples, as shown in (a) in Figure \ref{method}. Specifically, we employ the quantile function to establish a filtering threshold to select samples with higher uncertainty scores. Let $F$ denote the cumulative distribution function of the uncertainty scores $\text{AUV}$ for all training samples. The quantile function $F^{-1}:[0,1] \to \text{AUV}(Z_i)$ is then defined as:

\begin{equation}
\begin{aligned}
& F^{-1}(\tilde{p})=\inf\left\{\text{AUV}:\,\tilde{p}\leq F(\text{AUV}(Z_i))\right\}.
\end{aligned}
\end{equation}


\noindent where, $\tilde{p}$ is the specified quantile value and $F^{-1}(\tilde{p})$ represents the corresponding uncertainty threshold. We retain samples with uncertainty scores below the specified quantile of $\tilde{p}=95\%$ for model training:

\begin{equation}
\begin{aligned}
& \mathcal{D}^{*} = \left\{ \tilde{x}_i \mid \text{AUV} \leq F^{-1}(\tilde{p}) \right\}_{i=1}^{N}.
\end{aligned}
\end{equation}


Considering the challenges of ambiguity in manual annotations and high manual labeling costs, different image inputs can be selected. When $\hat{X}$ is used as input, the AUV directly reflects the inherent anatomical structural complexity of the image, the process that does not involve any annotation information. In addition, when $X$ is used as input, the AUV represents an intrinsic measure of annotation quality.


Due to medical tumor tasks that require primary focus on specific class regions, we also attempt to discard noisy samples based on each individual class. For each class $c \in {1,2,\ldots,C}$, we independently calculate the cumulative distribution function $F_c$ of the uncertainty scores for samples in that class and its quantile function $F^{-1}_c(\tilde{p})$, then retain samples in $\mathcal{D}^{*}$ that satisfy the following condition:

\begin{equation}
\begin{aligned}
& \mathcal{D}^{*} = \bigcup_{c=1}^{C} \left\{\tilde{x}_i \,\middle|\, \text{AUV} \leq F_{c}^{-1}(\tilde{p}),\; Z_{i}=c \right\}_{i=1}^{N_{c}}.
\end{aligned}
\end{equation}


\noindent where, $N_c$ denotes the number of samples belonging to class $c$. The efficacy of both aforementioned schemes is validated in the following experimental section.

\subsection{Dynamic Uncertainty-aware Optimization}


Existing studies typically model predictions following a Gaussian distribution parameterized by mean and variance, and design denoising strategies based on density estimation to improve predictive performance \cite{zhang2024one}. However, medical image segmentation is inherently a Bernoulli distribution problem, and strong distributional assumptions can introduce mismatch bias. Therefore, we directly leverage the semantic perception scale $\mathcal{S}$ of the model's predictive features to reveal its cognitive bias, as shown in Figure \ref{method}-b.


Assuming that only label noise exists in the samples, for each data pair $(x_i,\tilde{y}_i)$ where $\tilde{y}_i=y_i + \epsilon_i$, we add a learnable estimated noise $\hat{\epsilon}_i$. During the training of the model from scratch, the backbone network obtains the final prediction $Z_i = f(x_i)_{\theta_1}$ through the segmentation head. Simultaneously, independent of the backbone network, we add a noise estimation head $f(\cdot)_{\theta_2}$ to approximate the true noise and obtain clean data by label denoising: $\tilde{y}_i = y_i - \hat{\epsilon}_i \cdot f(x_i)_{\theta_2}$. Therefore, the segmentation loss based on binary cross-entropy and dice can be defined as:

\begin{equation}
\begin{aligned}
& \mathcal{L}_{\text{seg}} = \beta \cdot \mathcal{L}_{\text{Dice}}(z_i, \tilde{y}_i) + (1-\beta) \cdot \mathcal{L}_{\text{BCE}}(z_i, \tilde{y}_i),
\end{aligned}
\end{equation}


\noindent where $\beta$ is a hyperparameter that represents the weight of the binary loss function. To ensure the rationality and stability of noise estimation, we impose statistical constraints on estimated noise $\hat{\epsilon}_i$ to prevent divergence or invalid solutions during training. Furthermore, to dynamically adjust the model's attention during training and to achieve dynamic evaluation and weight assignment for uncertain classes, we utilize the value $\mathcal{S}(\cdot)$ to adaptively adjust the epistemic contribution of different classes in each iteration. Finally, the overall training loss can be defined as:

\begin{equation}
\begin{aligned}
& \mathcal{L}_{\text{total}} = \frac{1}{N}\sum_{i=1}^N \sum_{c=1}^C \frac{\mathcal{L}_{\text{seg}}(f(x_i)_{\theta_1}, y_i - \hat{\epsilon}_i \cdot f(x_i)_{\theta_2})}{1+\alpha \cdot \mathcal{S}(f(x_i|c)_{\theta_1})},\\
& \text{s.t.}\quad 
\frac{1}{N}\sum_{i=1}^{N} \hat{\epsilon}_{i} = 0, 
\quad 
\frac{1}{N}\sum_{i=1}^{N} \hat{\epsilon}_{i}^{2} = 1.
\end{aligned}
\end{equation}


\noindent Here, the hyperparameter $\alpha$ controls the contribution degree of $\mathcal{S}(\cdot)$. Subsequent experiments validate that the proposed dynamically uncertainty-aware optimization strategy can be deployed in a plug-and-play manner across different models, achieving consistent performance improvements.

\section{Experiments}

\subsection{Experiments Setup}


\textbf{Dataset selection.} Specifically, the LiTS dataset \cite{bilic2023liver} comprises contrast-enhanced 3D abdominal CT scans from 201 patients with hepatocellular carcinoma. These scans were manually annotated by experts to provide ground truth labels for the liver and liver tumors, and were subsequently divided into 131 training cases and 70 test cases. The TotalSegmentator dataset \cite{wasserthal2023totalsegmentator} represents the largest publicly available dataset for 3D medical image segmentation, containing 1,204 CT images covering 104 anatomical structures throughout the body. The dataset is partitioned into 1,082 training cases, 57 validation cases, and 65 test cases, comprising 27 organs, 59 bones, 10 muscles, and 8 blood vessels. The WORD dataset \cite{luo2022word} includes 120 comprehensively annotated abdominal CT scans with detailed pixel-level annotations covering 16 abdominal organs, along with sparse scribble-based annotations. The FeTA 2022 dataset \cite{payette2021automatic} provides 120 T2-weighted fetal brain MRI reconstructions sourced from two different medical institutions, accompanied by manual segmentation labels for seven distinct brain tissues. The KiTS23 dataset \cite{heller2023kits21} contains 489 3D CT images from kidney tumor patients, annotated with three semantic categories: kidney, kidney tumor, and renal cyst.

\begin{table*}[thbp]
\footnotesize
\centering
\caption{Performance of aleatoric uncertainty-aware data filtering strategies based on multiple medical foundation models and two baseline strategies across five medical segmentation datasets. Here, `95\%' indicates the removal of `5\%' of noisy samples.}
\vskip -0.1in
\begin{tabularx}{1.0\linewidth}{X<{\raggedright}|m{0.6cm}<{\centering}|m{0.9cm}<{\centering}m{0.9cm}<{\centering}m{0.9cm}<{\centering}m{0.9cm}<{\centering}m{0.9cm}<{\centering}m{0.9cm}<{\centering}m{0.9cm}<{\centering}m{0.9cm}<{\centering}m{0.9cm}<{\centering}m{0.9cm}<{\centering}m{0.5cm}<{\centering}}
\toprule
\multirow{1}*{Quantification} & \multirow{2}*{Data} & \multicolumn{2}{c}{LiTS} & \multicolumn{2}{c}{TotalSeg} & \multicolumn{2}{c}{WORD} & \multicolumn{2}{c}{FeTA} & \multicolumn{2}{c}{KiTS23} & \multirow{2}*{Mean}\\ 
\cmidrule(l){3-12}
Strategy & & Dice(\%) & mIoU(\%) & Dice(\%) & mIoU(\%) & Dice(\%) & mIoU(\%) & Dice(\%) & mIoU(\%) & Dice(\%) & mIoU(\%) & \\
\midrule
\cellcolor{lightgray} Baseline & \cellcolor{lightgray} 100\% & \cellcolor{lightgray}82.96 & \cellcolor{lightgray}73.24 & \cellcolor{lightgray}80.34 & \cellcolor{lightgray}72.71 & \cellcolor{lightgray}79.03 & \cellcolor{lightgray}69.85 & \cellcolor{lightgray}84.30 & \cellcolor{lightgray}74.63 & \cellcolor{lightgray}70.15 & \cellcolor{lightgray}63.76 & \cellcolor{lightgray}75.10\\
\midrule
\multicolumn{13}{c}{\textbf{Quantifying AUV via Data Variance.}}\\
\midrule
\multirow{2}{*}{Variance \cite{zhang2024one}} & 95\%  & 83.00{\scriptsize \textcolor{green}{$\uparrow$}} & 73.27{\scriptsize \textcolor{green}{$\uparrow$}} & 80.29{\scriptsize \textcolor{red}{$\downarrow$}} & 72.66{\scriptsize \textcolor{red}{$\downarrow$}} & 81.43{\scriptsize \textcolor{green}{$\uparrow$}} & 71.99{\scriptsize \textcolor{green}{$\uparrow$}} & 84.50{\scriptsize \textcolor{green}{$\uparrow$}} & 74.76{\scriptsize \textcolor{green}{$\uparrow$}} & 68.05{\scriptsize \textcolor{red}{$\downarrow$}} & 61.97{\scriptsize \textcolor{red}{$\downarrow$}}& 75.19{\scriptsize \textcolor{green}{$\uparrow$}}\\
 & 90\%  & 81.72{\scriptsize \textcolor{red}{$\downarrow$}} & 72.15{\scriptsize \textcolor{red}{$\downarrow$}} & 80.37{\scriptsize \textcolor{green}{$\uparrow$}} & 72.93{\scriptsize \textcolor{green}{$\uparrow$}} & 81.67{\scriptsize \textcolor{green}{$\uparrow$}} & 72.15{\scriptsize \textcolor{green}{$\uparrow$}} & 84.52{\scriptsize \textcolor{green}{$\uparrow$}} & 74.78{\scriptsize \textcolor{green}{$\uparrow$}} & 69.96{\scriptsize \textcolor{red}{$\downarrow$}} & 63.66{\scriptsize \textcolor{red}{$\downarrow$}}& 75.39{\scriptsize \textcolor{green}{$\uparrow$}}\\
\midrule
\multicolumn{13}{c}{\textbf{Quantifying AUV via Task-specific Models.}}\\
\midrule
\multirow{2}{*}{nnU-Net \cite{isensee2021nnu}} & 95\% & 83.12{\scriptsize \textcolor{green}{$\uparrow$}} & 73.38{\scriptsize \textcolor{green}{$\uparrow$}} & 80.33{\scriptsize \textcolor{red}{$\downarrow$}} & 72.59{\scriptsize \textcolor{red}{$\downarrow$}} & 79.90{\scriptsize \textcolor{green}{$\uparrow$}} & 70.74{\scriptsize \textcolor{green}{$\uparrow$}} & 84.58{\scriptsize \textcolor{green}{$\uparrow$}} & 74.85{\scriptsize \textcolor{green}{$\uparrow$}} & 70.09{\scriptsize \textcolor{red}{$\downarrow$}}& 63.71{\scriptsize \textcolor{red}{$\downarrow$}}& 75.33{\scriptsize \textcolor{green}{$\uparrow$}}\\
& 90\% & 83.47{\scriptsize \textcolor{green}{$\uparrow$}} & 73.69{\scriptsize \textcolor{green}{$\uparrow$}} & 79.86{\scriptsize \textcolor{red}{$\downarrow$}} & 72.14{\scriptsize \textcolor{red}{$\downarrow$}} & 80.30{\scriptsize \textcolor{green}{$\uparrow$}} & 71.01{\scriptsize \textcolor{green}{$\uparrow$}} & 84.64{\scriptsize \textcolor{green}{$\uparrow$}} & 74.92{\scriptsize \textcolor{green}{$\uparrow$}} & 69.02{\scriptsize \textcolor{red}{$\downarrow$}}& 62.66{\scriptsize \textcolor{red}{$\downarrow$}}& 75.17{\scriptsize \textcolor{green}{$\uparrow$}}\\
\midrule
\multicolumn{13}{c}{\textbf{Quantifying AUV via Medical Visual Foundation Models.}}\\
\midrule
 \multirow{2}{*}{CLIP-Driven \cite{liu2023clip}} & 95\%  & 82.18{\scriptsize \textcolor{red}{$\downarrow$}} & 72.56{\scriptsize \textcolor{red}{$\downarrow$}} & 80.12{\scriptsize \textcolor{red}{$\downarrow$}} & 72.57{\scriptsize \textcolor{red}{$\downarrow$}} & 80.14{\scriptsize \textcolor{green}{$\uparrow$}} & 70.84{\scriptsize \textcolor{green}{$\uparrow$}} & 84.59{\scriptsize \textcolor{green}{$\uparrow$}} & 74.85{\scriptsize \textcolor{green}{$\uparrow$}} & 69.36{\scriptsize \textcolor{red}{$\downarrow$}}& 63.12{\scriptsize \textcolor{red}{$\downarrow$}}& 75.03{\scriptsize \textcolor{red}{$\downarrow$}}\\
 & 90\%  & 82.21{\scriptsize \textcolor{red}{$\downarrow$}} & 72.59{\scriptsize \textcolor{red}{$\downarrow$}} & 79.53{\scriptsize \textcolor{red}{$\downarrow$}} & 71.95{\scriptsize \textcolor{red}{$\downarrow$}} & 81.15{\scriptsize \textcolor{green}{$\uparrow$}} & 71.69{\scriptsize \textcolor{green}{$\uparrow$}} & 84.51{\scriptsize \textcolor{green}{$\uparrow$}} & 74.71{\scriptsize \textcolor{green}{$\uparrow$}} & 69.84{\scriptsize \textcolor{red}{$\downarrow$}}& 63.57{\scriptsize \textcolor{red}{$\downarrow$}}& 75.18{\scriptsize \textcolor{green}{$\uparrow$}}\\
 \midrule
\multirow{2}{*}{SegVol \cite{du2024segvol}} & 95\%  & 84.51{\scriptsize \textcolor{green}{$\uparrow$}} & 74.62{\scriptsize \textcolor{green}{$\uparrow$}} & 80.56{\scriptsize \textcolor{green}{$\uparrow$}} & 72.73{\scriptsize \textcolor{green}{$\uparrow$}} & 80.12{\scriptsize \textcolor{green}{$\uparrow$}} & 70.95{\scriptsize \textcolor{green}{$\uparrow$}} & 84.66{\scriptsize \textcolor{green}{$\uparrow$}} & 74.97{\scriptsize \textcolor{green}{$\uparrow$}} & 70.31{\scriptsize \textcolor{green}{$\uparrow$}}& 63.95{\scriptsize \textcolor{green}{$\uparrow$}}& 75.73{\scriptsize \textcolor{green}{$\uparrow$}}\\
 & 90\%  & 83.73{\scriptsize \textcolor{green}{$\uparrow$}} & 74.03{\scriptsize \textcolor{green}{$\uparrow$}} & 81.59{\scriptsize \textcolor{green}{$\uparrow$}} & 74.03{\scriptsize \textcolor{green}{$\uparrow$}} & 81.52{\scriptsize \textcolor{green}{$\uparrow$}} & 72.01{\scriptsize \textcolor{green}{$\uparrow$}} & \textbf{84.76}{\scriptsize \textcolor{green}{$\uparrow$}} & \textbf{74.97}{\scriptsize \textcolor{green}{$\uparrow$}} & 70.52{\scriptsize \textcolor{green}{$\uparrow$}}& 63.98{\scriptsize \textcolor{green}{$\uparrow$}}& 76.11{\scriptsize \textcolor{green}{$\uparrow$}}\\
\midrule
\multirow{2}{*}{MedSAM2 \cite{ma2025medsam2}} & 95\%  & 83.31{\scriptsize \textcolor{green}{$\uparrow$}} & 73.55{\scriptsize \textcolor{green}{$\uparrow$}} & 81.31{\scriptsize \textcolor{green}{$\uparrow$}} & 73.76{\scriptsize \textcolor{green}{$\uparrow$}} & 80.27{\scriptsize \textcolor{green}{$\uparrow$}} & 71.04{\scriptsize \textcolor{green}{$\uparrow$}} & 84.72{\scriptsize \textcolor{green}{$\uparrow$}} & 74.99{\scriptsize \textcolor{green}{$\uparrow$}} & 70.43{\scriptsize \textcolor{green}{$\uparrow$}}& 63.95{\scriptsize \textcolor{green}{$\uparrow$}}& 75.73{\scriptsize \textcolor{green}{$\uparrow$}}\\
 & 90\%  & \textbf{85.45}{\scriptsize \textcolor{green}{$\uparrow$}} & \textbf{75.47}{\scriptsize \textcolor{green}{$\uparrow$}} & \textbf{81.65}{\scriptsize \textcolor{green}{$\uparrow$}} & \textbf{73.99}{\scriptsize \textcolor{green}{$\uparrow$}} & \textbf{81.74}{\scriptsize \textcolor{green}{$\uparrow$}} & \textbf{72.18}{\scriptsize \textcolor{green}{$\uparrow$}} & 84.59{\scriptsize \textcolor{green}{$\uparrow$}} & 74.82{\scriptsize \textcolor{green}{$\uparrow$}} & \textbf{70.57}{\scriptsize \textcolor{green}{$\uparrow$}}& \textbf{63.99}{\scriptsize \textcolor{green}{$\uparrow$}}& \textbf{76.45}{\scriptsize \textcolor{green}{$\uparrow$}}\\
\bottomrule
\end{tabularx}
\label{cp_1}
\vskip -0.10in
\end{table*}

\begin{table*}[thbp]
\footnotesize
\centering
\caption{Performance of the Dynamic Uncertainty-aware Optimization (DUO) strategy on three baseline networks.}
\vskip -0.1in
\begin{tabularx}{1.0\linewidth}{X<{\raggedright}|m{0.5cm}<{\centering}|m{0.9cm}<{\centering}m{0.9cm}<{\centering}|m{0.9cm}<{\centering}m{0.9cm}<{\centering}|m{0.9cm}<{\centering}m{0.9cm}<{\centering}|m{0.9cm}<{\centering}m{0.9cm}<{\centering}|m{0.9cm}<{\centering}m{0.9cm}<{\centering}|m{0.55cm}<{\centering}}
\toprule
\multirow{2}*{Method} & \multirow{2}*{DUO} & \multicolumn{2}{c}{LiTS} & \multicolumn{2}{c}{TotalSeg} & \multicolumn{2}{c}{WORD} & \multicolumn{2}{c}{FeTA} & \multicolumn{2}{c}{KiTS23} & \multirow{2}*{Mean}\\ 
\cmidrule(l){3-12}
& & Dice(\%) & mIoU(\%) & Dice(\%) & mIoU(\%) & Dice(\%) & mIoU(\%) & Dice(\%) & mIoU(\%) &Dice(\%) & mIoU(\%) &\\
\midrule
\multirow{2}{*}{nnU-Net \cite{isensee2021nnu}} & $\times$  & \cellcolor{lightgray}82.96 & \cellcolor{lightgray}73.24 & \cellcolor{lightgray}80.34 & \cellcolor{lightgray}72.71 & \cellcolor{lightgray}79.03 & \cellcolor{lightgray}69.85 & \cellcolor{lightgray}84.30 & \cellcolor{lightgray}74.63 & \cellcolor{lightgray}70.15 & \cellcolor{lightgray}63.76 & \cellcolor{lightgray}75.10 \\
& $\checkmark$ & 84.13{\scriptsize \textcolor{green}{$\uparrow$}} & 74.73{\scriptsize \textcolor{green}{$\uparrow$}} & 80.82{\scriptsize \textcolor{green}{$\uparrow$}} & 73.14{\scriptsize \textcolor{green}{$\uparrow$}} & 79.63{\scriptsize \textcolor{green}{$\uparrow$}} & 70.38{\scriptsize \textcolor{green}{$\uparrow$}} & 84.63{\scriptsize \textcolor{green}{$\uparrow$}} & 74.91{\scriptsize \textcolor{green}{$\uparrow$}} & 70.51{\textcolor{green}{$\uparrow$}}& 63.97{\textcolor{green}{$\uparrow$}} & 75.69{\textcolor{green}{$\uparrow$}}\\
 \midrule
\multirow{2}{*}{Swin-UNETR \cite{he2023swinunetr}} & $\times$  & \cellcolor{lightgray}80.54 & \cellcolor{lightgray}71.10 & \cellcolor{lightgray}77.38 & \cellcolor{lightgray}70.03 & \cellcolor{lightgray}77.26 & \cellcolor{lightgray}68.26 & \cellcolor{lightgray}84.15 & \cellcolor{lightgray}74.48 & \cellcolor{lightgray}66.89& \cellcolor{lightgray}60.79& \cellcolor{lightgray}73.09\\
 & $\checkmark$ & 81.25{\scriptsize \textcolor{green}{$\uparrow$}} & 71.72{\scriptsize \textcolor{green}{$\uparrow$}} & 77.94{\scriptsize \textcolor{green}{$\uparrow$}} & 70.51{\scriptsize \textcolor{green}{$\uparrow$}} & 77.90{\scriptsize \textcolor{green}{$\uparrow$}} & 68.85{\scriptsize \textcolor{green}{$\uparrow$}} & 84.53{\scriptsize \textcolor{green}{$\uparrow$}} & 74.82{\scriptsize \textcolor{green}{$\uparrow$}} & 67.57{\textcolor{green}{$\uparrow$}}& 61.32{\textcolor{green}{$\uparrow$}} & 73.64{\textcolor{green}{$\uparrow$}}\\
\midrule
 \multirow{2}{*}{U-Mamba \cite{ma2024u}} & $\times$  & \cellcolor{lightgray}81.24 & \cellcolor{lightgray}71.72 & \cellcolor{lightgray}79.27 & \cellcolor{lightgray}71.74 & \cellcolor{lightgray}78.93 & \cellcolor{lightgray}69.76 & \cellcolor{lightgray}84.42 & \cellcolor{lightgray}74.72 & \cellcolor{lightgray}67.85& \cellcolor{lightgray}61.67& \cellcolor{lightgray}74.13\\
 & $\checkmark$ & 81.92{\scriptsize \textcolor{green}{$\uparrow$}} & 72.43{\scriptsize \textcolor{green}{$\uparrow$}} & 79.85{\scriptsize \textcolor{green}{$\uparrow$}} & 72.27{\scriptsize \textcolor{green}{$\uparrow$}} & 79.25{\scriptsize \textcolor{green}{$\uparrow$}} & 70.04{\scriptsize \textcolor{green}{$\uparrow$}} & 84.77{\scriptsize \textcolor{green}{$\uparrow$}} & 75.03{\scriptsize \textcolor{green}{$\uparrow$}} & 68.69{\textcolor{green}{$\uparrow$}} & 62.33{\textcolor{green}{$\uparrow$}}& 74.66{\textcolor{green}{$\uparrow$}}\\
\bottomrule
\end{tabularx}
\label{cp_2}
\vskip -0.10in
\end{table*}


\textbf{Evaluation Metric.} We adopt the Dice score and the mean Intersection over Union (mIoU) as evaluation metrics. To fairness, we report the inference results of the model on the test set rather than the training results for the best epoch.



\textbf{Implementation Details.} To ensure fair and comprehensive comparisons, all experiments maintain consistent configurations. We uniformly apply nnU-Net's preprocessing for data normalization and resampling. Models are trained from scratch for 100 epochs using SGD with an initial learning rate of 0.01, updated per epoch via a Poly policy. Training uses $96\times96\times96$ patches with a batch size of 2. Each dataset is split with 20\% reserved for testing. All implementations use Python 3.10, PyTorch 2.0.0, and a single RTX 4090 GPU.

\subsection{Performance on Aleatoric Uncertainty-aware Data Filtering}



To ensure experimental fairness, we consistently employ nnU-Net \cite{isensee2021nnu} as the baseline network and use normalized images as input, with the mean aleatoric uncertainty across all classes as the AUV. The baseline is established by training the network on 100\% of the training set and evaluating predictions on a fixed test set. We compare classical strategies that use data variance as an uncertainty metric to eliminate noisy data and further evaluate the effectiveness of task-specific models versus foundation models in quantifying AUV, as summarized in Table \ref{cp_1}. The results demonstrate that directly computed data features still fail to accurately represent data uncertainty in practice. Furthermore, the results indicate that using task-specific models as feature extractors yields only minimal performance improvements, potentially due to feature overfitting. Additionally, the appendix discusses how directly using the model's predicted Dice score as the AUV leads to more severe overfitting issues. Finally, the experiments validate the effectiveness of employing visual foundation models as feature extractors, where MedSAM2 \cite{ma2025medsam2} achieves the best performance, while not all foundation models produce positive improvements in these segmentation tasks.

\subsection{Performance on Dynamic Uncertainty-aware Optimization}


Independent of the aforementioned data filtering, the Dynamic Uncertainty-Aware Optimization (DUO) strategy demonstrates flexible applicability across diverse medical segmentation scenarios. Therefore, we validate the approach on the same five datasets using three distinct backbone architectures including CNN, Transformer and Mamba. All critical experimental configurations, including data pre-processing, post-processing and testing protocols, are maintained rigorously consistent. As delineated in Table \ref{cp_2}, the results substantiate that the DUO strategy achieves stable and widespread performance improvements across different backbone networks with minimal computational cost. This enhancement is visually corroborated in Figure \ref{visual_sps}, where the integration of DUO leads to more precise segmentation predictions, especially at the target boundaries, compared to the baseline. Additionally, We also provide comparative experiments between DUO and the static re-weighting method in the Appendix.

\begin{figure}[tbp]
\centering
\includegraphics[width=0.99\linewidth]{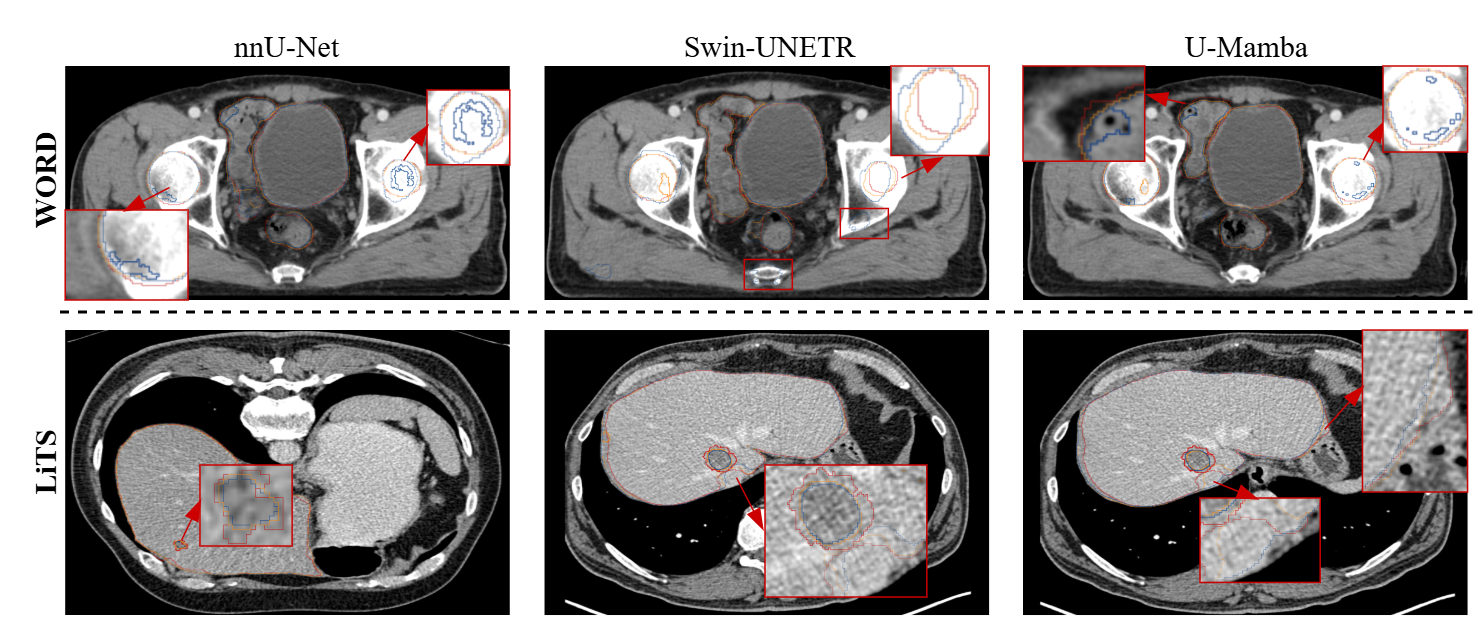}
\vskip -0.1in
\caption{The comparison of segmentation boundaries from three baseline networks on tumor (LiTS) and multi-organ (WORD) dataset. Ground truth (red) is compared against the baseline (blue) and the DUO strategy (orange).}
\label{visual_sps}
\vskip -0.1in
\end{figure}

\begin{table}[tbp]
\footnotesize
\centering
\caption{Comparison of different uncertainty quantification methods on tumor segmentation (Dice \%) in two medical applications of the LiTS dataset.}
\begin{tabularx}{\columnwidth}{X<{\centering}|m{1.0cm}<{\centering}m{1.2cm}<{\centering}|m{1.0cm}<{\centering}m{1.2cm}<{\centering}} 
\toprule
\multirow{2}*{Method} & \multicolumn{2}{c}{90 \% data} & \multicolumn{2}{c}{+DUO}  \\
\cmidrule(l){2-5}
 & Liver & Tumor & Liver & Tumor\\
\midrule
Fisher \cite{deng2023uncertainty}  & 96.58 & 70.17$_{\textcolor{green}{+0.86}}$& 96.59 & 70.01$_{\textcolor{green}{+0.70}}$\\
MD \cite{cui2024exploring}  & 96.56 & 69.78$_{\textcolor{green}{+0.47}}$& 96.57 & 69.28$_{\textcolor{red}{-0.03}}$\\
EAOA \cite{zong2025rethinking} & 96.61 & 70.41$_{\textcolor{green}{+1.10}}$ & $-$ & $-$ \\
\midrule
$\mathcal{S}(\cdot)$  & 96.61 & 74.28$_{\textcolor{green}{+4.97}}$ & 96.60 & 71.66$_{\textcolor{green}{+2.35}}$\\ 
\bottomrule
\end{tabularx}
\label{cp_sps}
\vskip -0.2in
\end{table}

\subsection{Ablation Studies and Analysis}


\textbf{Effectiveness of semantic perception scale.} The core of the proposed AUV lies in using the semantic perception scale $\mathcal{S}(\cdot)$ of quantized visual foundation models to measure the uncertainty of the image. Therefore, the experiment compares multiple quantification methods that leverage model’s perceptual features, including strategies based on the Mahalanobis Distance (MD) \cite{cui2024exploring}, Fisher metric \cite{deng2023uncertainty} and EAOA \cite{zong2025rethinking}. The experimental protocols and settings remain consistent with Method Figure \ref{method}, using nnU-Net as the baseline network for training and testing on the LiTS dataset. Since EAOA is designed for sample-level AUV quantification and cannot perform semantic perception scale calculations for individual classes, the comparative analysis is thus limited to its performance in data filtering strategies. Table \ref{cp_sps} presents a comprehensive comparison of these methods applied to uncertainty-aware data filtering and dynamic uncertainty-aware optimization strategy. The results demonstrate that the proposed method achieves advanced performance and flexible adaptability in both applications. Compared to alternative quantification schemes, this singular value energy-based quantification method derived from feature matrices proves better suited for medical imaging tasks.

\begin{figure}[tbp]
\centering
\includegraphics[width=0.99\linewidth]{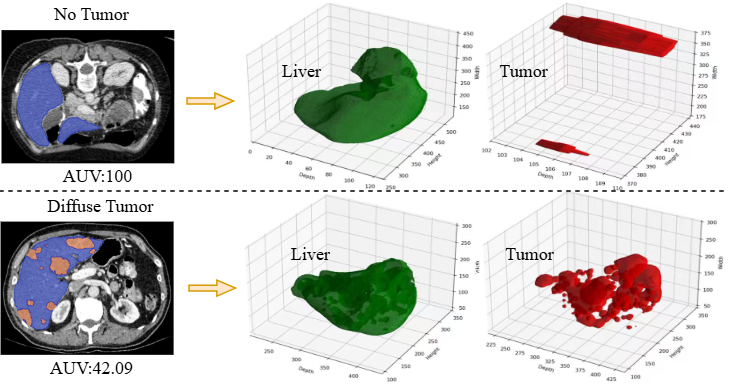}
\vskip -0.1in
\caption{The correspondence between the feature vectors $Z$ of the liver and tumors predicted by MedSAM2 and the aleatoric uncertainty values is visualized through t-SNE.}
\label{visual_sps}
\vskip -0.2in
\end{figure}

\begin{table}[tbp]
\footnotesize
\centering
\caption{Correlation analysis between quantified values from different methods and predicted Dice scores}
\begin{tabularx}{\columnwidth}{X<{\centering}|m{1.2cm}<{\centering}m{0.9cm}<{\centering}|m{1.2cm}<{\centering}m{0.9cm}<{\centering}}
\toprule
Properties & Pearson & P-value & Spearman & P-value \\
\midrule
\midrule
Fisher \cite{deng2023uncertainty} & 0.4402 & 0.0880 & 0.6623 & 0.0052\\
MD \cite{cui2024exploring} & 0.3620 & 0.1683 & 0.4618 & 0.0718\\
\midrule
$\mathcal{S}(\cdot)$ & 0.6267 & 0.0094 & 0.7471 & 0.0009\\
\bottomrule
\end{tabularx}
\label{Correlation}
\vskip -0.1in
\end{table}

\begin{table}[tbp]
\footnotesize
\centering
\caption{The impact of different data filtering strategies on tumor segmentation performance in LiTS.}
\begin{tabularx}{\columnwidth}{X<{\centering}|m{1.2cm}<{\centering}m{1.2cm}<{\centering}|m{1.2cm}<{\centering}m{1.2cm}<{\centering}} 
\toprule
\multirow{2}*{strategy} & \multicolumn{2}{c}{90 \% data} & \multicolumn{2}{c}{95 \% data}  \\
\cmidrule(l){2-5}
 & Dice (\%) & mIoU (\%) & Dice (\%) & mIoU (\%)\\
\midrule
(a) & 73.35$_{\textcolor{green}{+4.04}}$ & 63.44$_{\textcolor{green}{+2.99}}$ & 69.47$_{\textcolor{green}{+0.16}}$ & 59.43$_{\textcolor{green}{+0.12}}$\\
(b) & 72.38$_{\textcolor{green}{+2.53}}$ & 62.32$_{\textcolor{green}{+1.87}}$ & 69.76$_{\textcolor{green}{+0.45}}$ & 59.73$_{\textcolor{green}{+0.42}}$\\
\midrule
(c) & 74.28$_{\textcolor{green}{+4.97}}$ & 63.69$_{\textcolor{green}{+3.24}}$ & 70.03$_{\textcolor{green}{+0.72}}$ & 59.88$_{\textcolor{green}{+0.57}}$\\
(d) & 73.74$_{\textcolor{green}{+4.43}}$ & 63.29$_{\textcolor{green}{+2.84}}$ & 70.26$_{\textcolor{green}{+0.95}}$ & 60.07$_{\textcolor{green}{+0.76}}$\\
\bottomrule
\end{tabularx}
\label{AUDF}
\vskip -0.2in
\end{table}


The quantification of model perceptual features does not rely on annotations, but whether its expressed physical meaning aligns with actual prediction trends warrants investigation. Therefore, we further explore the fundamental differences among these quantification schemes through correlation analysis of model predictions, as shown in Table \ref{Correlation}. The results indicate that the proposed method demonstrates the strongest correlation with the model's predicted Dice scores for each class, thereby explaining its superior performance in both data filtering and optimization strategies for model training. Additionally, Figure \ref{visual_sps} visualizes the feature vectors $Z$ of the liver and tumors predicted by MedSAM2, revealing that the model exhibits misplaced attention to tumor regions in noise samples.

\begin{figure}[tbp]
\centering
\includegraphics[width=0.99\linewidth]{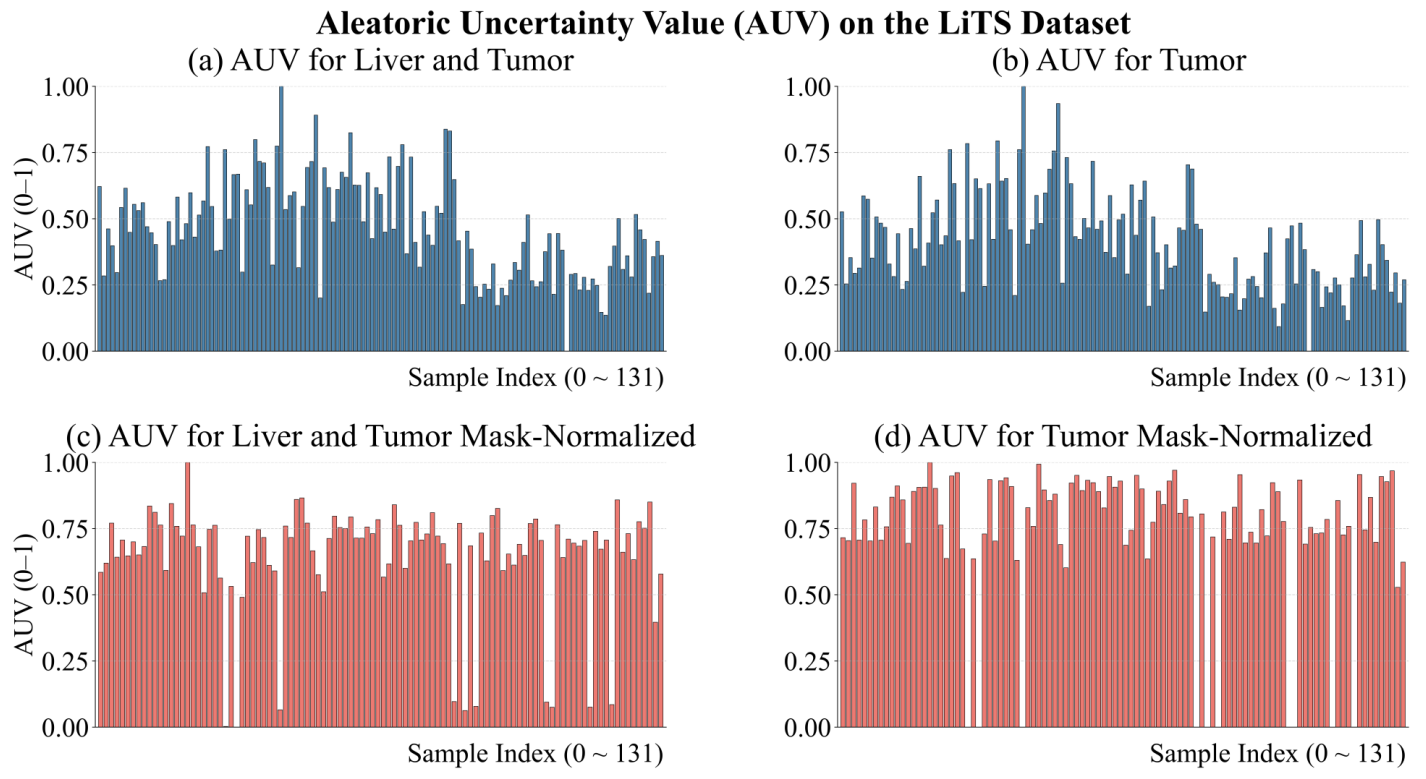}
\vskip -0.1in
\caption{The histograms of AUV based on two different image inputs for both entire and specific class analyses in the LiTS dataset.}
\label{auv_distribution}
\vskip -0.2in
\end{figure}

\textbf{Effectiveness of aleatoric uncertainty-aware data filtering mechanisms.} 
All experiments consistently used nnU-Net as the baseline network and employed MedSAM2 as the foundation model for computing the AUV. Table \ref{AUDF} presents the performance improvements achieved by aleatoric uncertainty-aware data filtering mechanisms across four practical scenarios, differentiated by the presence or absence of annotations and the selection of specific classes. These four filtering mechanisms correspond to the four distinct AUV distributions illustrated in Figure \ref{auv_distribution}. The results demonstrate that mechanisms (c) and (d) significantly outperform mechanisms (a) and (b), indicating that images normalized with foreground mask regions are more suitable for assessing aleatoric uncertainty. Although this strategy requires annotation information, the experimental results also confirm the effectiveness of evaluating aleatoric uncertainty directly using raw images. Furthermore, the performance difference between the data filtering applied specifically to tumor class and all classes is minimal, which may be attributed to the relative simplicity of the liver class in the LiTS dataset, making it difficult to exert a substantial influence on the AUV distribution. The appendix provides comprehensive discussions on data filtering ratios and additional experiments with more backbone networks.

\begin{figure}[tbp]
\centering
\includegraphics[width=0.99\linewidth]{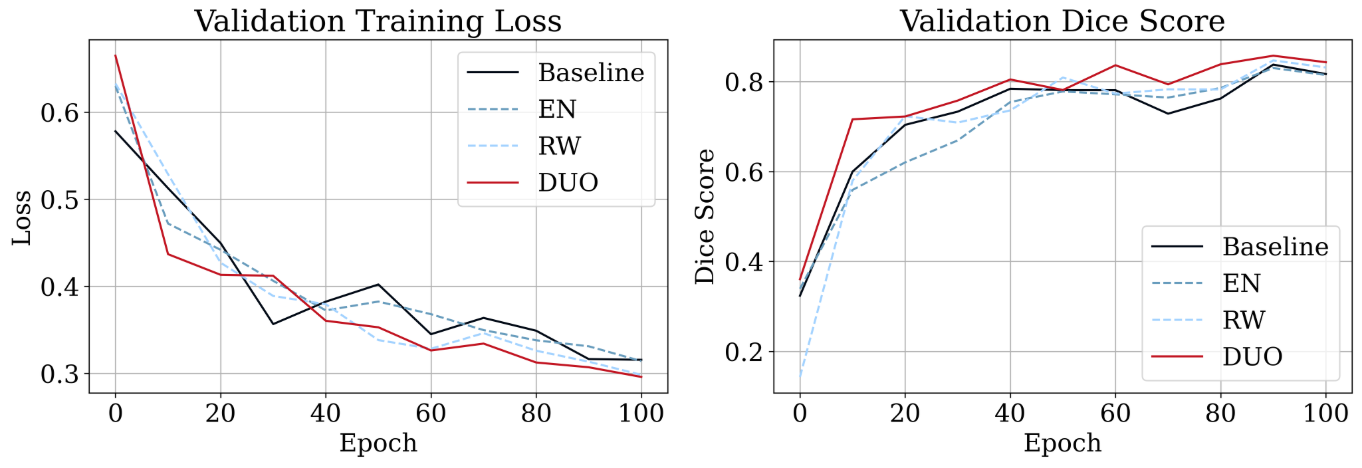}
\vskip -0.1in
\caption{Validation Loss and local Dice score curves of the dynamic uncertainty-aware optimization strategy during training.}
\label{Training_curve}
\vskip -0.2in
\end{figure}

\textbf{Effectiveness of dynamic uncertainty-aware optimization.} 
The experiments use the combination of Dice and BCE loss as the baseline, conducting ablation studies on key components of the dynamic uncertainty-aware optimization strategy: Estimated Noise in annotations (EN) and the Re-Weighting (RW) based on semantic perception scale. The EN component transforms label representations from binary distributions into continuous values, thereby no longer treating annotations as absolutely correct, but rather incorporating a denoising process. Moreover, the RW mechanism prioritizes the allocation of limited cognitive resources to challenging samples with ambiguous classifications and complex features, thus achieving a more efficient gradient distribution across the overall data distribution and promoting a more balanced convergence process. Table \ref{ab_duo} demonstrates the performance improvement achieved by these two design components, thus validating the rationality of our design. Furthermore, Figure \ref{Training_curve} illustrates the validation loss curves and the validation Dice score during the training process of the aforementioned experiments. The results demonstrate that DUO exhibits enhanced stability and effectiveness compared to baseline loss.

\begin{table}[tbp]
\footnotesize
\centering
\caption{Ablation study on the design of dynamic uncertainty-Aware optimization in the LiTS Dataset.}
\begin{tabularx}{\columnwidth}{X<{\centering}m{0.8cm}<{\centering}m{1.2cm}<{\centering}m{1.3cm}<{\centering}m{1.0cm}<{\centering}m{1.0cm}<{\centering}}
\toprule
EN & RW & Params(M) & Memory(M) & Liver & Tumor \\
\midrule
$\times$ & $\times$ & \textbf{30.80} & \textbf{6105} & 96.61 & 69.31 \\
\midrule
$\checkmark$ & $\times$ & 30.80 & 6105 & 96.60 & 69.87 \\
$\times$ & $\checkmark$ & 30.81 & 6287 & 96.59 & 71.37 \\
$\checkmark$ & $\checkmark$ & 30.81 & 6287 & \textbf{96.61} & \textbf{71.66} \\
\bottomrule
\end{tabularx}
\label{ab_duo}
\end{table}

\vspace{-0.2em}
\section{Conclusions}


This work addresses the pervasive issue of intrinsic randomness in medical image segmentation by proposing a general uncertainty quantification and utilization framework based on visual foundation models. By analyzing the singular value energy structure of feature vectors from foundation models, the method constructs semantically-aware scales to characterize sample difficulty and stochastic uncertainty. Furthermore, it designs a stochastic uncertainty-aware data filtering strategy and a dynamic uncertainty-aware optimization mechanism, enhancing model robustness to noisy samples and ambiguous boundaries. Extensive experiments demonstrate that the proposed approach achieves consistent performance gains across multiple modalities, tasks, and network architectures, validating the application potential and universal value of stochastic uncertainty in medical image analysis. Future work may extend to weakly supervised learning, federated learning, and dynamic data quality management in clinical settings to build more trustworthy, robust, and adaptive medical AI systems.

{
    \small
    \bibliographystyle{ieeenat_fullname}
    \bibliography{main}
}


\end{document}